\documentclass[letterpaper, 10 pt, journal, twoside]{IEEEtran}

\IEEEoverridecommandlockouts                              

\usepackage{amssymb}
\usepackage{booktabs}
\usepackage{graphicx}
\usepackage{multirow}
\usepackage{bbding}
\usepackage{url}
\usepackage{amsmath}
\usepackage{cite}
\usepackage{array}
\usepackage{xcolor}
\definecolor{darkgreen}{rgb}{0.0, 0.5, 0.0}
\newcolumntype{C}[1]{>{\centering\arraybackslash}p{#1}}
\newcommand\thickhline{\noalign{\hrule height 1.0pt}}
\usepackage[colorlinks,linkcolor=red,anchorcolor=blue,citecolor=green]{hyperref}

\newcommand{\orcid}[1]{\href{https://orcid.org/#1}{\includegraphics[width=10pt]{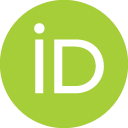}}}

\title{4D-CS: Exploiting Cluster Prior for 4D Spatio-Temporal LiDAR Semantic Segmentation
}

\author{Jiexi Zhong\orcid{0009-0000-3388-7865}, Zhiheng Li\orcid{0000-0002-1477-2066}, Yubo Cui\orcid{0000-0001-5302-0484}, Zheng Fang*\orcid{0000-0003-3887-3141}
\thanks{Manuscript received: July 22, 2024; Revised: October 16, 2024; Accepted: November 14, 2024.}
\thanks{This paper was recommended for publication by Editor Cesar Cadena Lerma upon evaluation of the Associate Editor and Reviewers' comments. This work was supported in part by the National Natural Science Foundation of China under Grants 62073066, in part by the Fundamental Research Funds for the Central Universities under Grant N2226001, and in part by 111 Project under Grant B16009. (Corresponding author: Zheng Fang)}
\thanks{The authors are all with the Faculty of Robot Science and Engineering, Northeastern University, Shenyang 110819, China. Jiexi Zhong and Zheng Fang are also with the National Frontiers Science Center for Industrial Intelligence and Systems Optimization, Northeastern University, Shenyang 110819, China and also with Key Laboratory of Data Analytics and Optimization for Smart Industry, Ministry of Education, Northeastern University, Shenyang 110819, China. (e-mail: fangzheng@mail.neu.edu.cn)}
\thanks{Digital Object Identifier (DOI): see top of this page.}
}

\begin{document}

\markboth{IEEE Robotics and Automation Letters. Preprint Version. Accepted November~2024}
{Zhong \MakeLowercase{\textit{et al.}}: 4D-CS: Exploiting Cluster Prior for 4D Spatio-Temporal LiDAR Semantic Segmentation} 

\maketitle

\begin{abstract}
Semantic segmentation of LiDAR points has significant value for autonomous driving and mobile robot systems. Most approaches explore spatio-temporal information of multi-scan to identify the semantic classes and motion states for each point.
However, these methods often overlook the segmentation consistency in space and time, which may result in point clouds within the same object being predicted as different categories.
To handle this issue, our core idea is to generate cluster labels across multiple frames that can reflect the complete spatial structure and temporal information of objects. These labels serve as explicit guidance for our dual-branch network, 4D-CS, which integrates point-based and cluster-based branches to enable more consistent segmentation.
Specifically, in the point-based branch, we leverage historical knowledge to enrich the current feature through temporal fusion on multiple views. 
In the cluster-based branch, we propose a new strategy to produce cluster labels of foreground objects and apply them to gather point-wise information to derive cluster features.  We then merge neighboring clusters across multiple scans to restore missing features due to occlusion.
Finally, in the point-cluster fusion stage, we adaptively fuse the information from the two branches to optimize segmentation results.
Extensive experiments confirm the effectiveness of the proposed method, and we achieve state-of-the-art results on the multi-scan semantic and moving object segmentation on SemanticKITTI and nuScenes datasets.
The code will be available at \url{https://github.com/NEU-REAL/4D-CS.git}.
\end{abstract}


\section{Introduction}

\IEEEPARstart{S}{emantic} segmentation of LiDAR points is a crucial task in autonomous driving and robotics, aiming at predicting the semantic categories of each point.
It is of great significance for the downstream tasks, including the semantic mapping~\cite{suma++} and long-term autonomous navigation~\cite{navigation}.

In recent years, several approaches~\cite{kpconv,pointnet++,PointNet,cylinder3d,sphereformer,cenet,rethinkingRange} have attempted semantic segmentation on a single LiDAR frame. However, these one-by-one segmentation algorithms ignore some useful temporal knowledge, especially the distinct and complementary observations of objects from past moments, making it difficult to handle cases with occlusion and sparse points. Moreover, due to separating each frame independently, these methods cannot distinguish the motion state of objects in a LiDAR sequence, leading to a ghost effect during mapping.

To overcome the above limitations, several methods adopt multi-scan LiDAR points to restore the complete appearance of objects~\cite{svqnet} or exploit spatio-temporal features to improve scene perception ability~\cite{templidarseg,metarangeseg,memoryseg}. 
In addition, they explore the potential moving information from a LiDAR sequence to identify the motion states of objects.
For example, MemorySeg~\cite{memoryseg} recurrently updates a memory feature to compensate for information loss caused by occlusion in the current frame.
SVQNet~\cite{svqnet} aggregates information from adjacent historical points for local feature encoding and selects temporal context to complete invisible geometry, leading to promising results.

\begin{figure}[t]
\centering
\includegraphics[width=8.0cm, height=5.8cm]{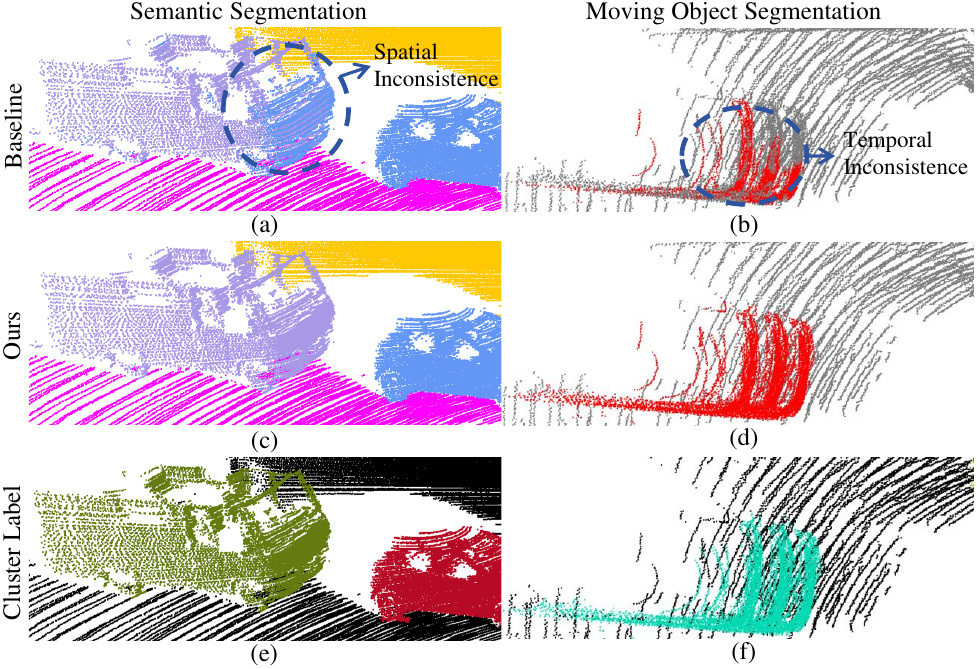}
\vspace{-0.1in}
\caption{The comparison of the baseline (WaffleIron~\cite{waffle}) with our proposed method on SemanticKITTI. For both methods, subfigures (a) and (c) display semantic segmentation, while subfigures (b) and (d) illustrate moving object segmentation. Subfigures (e) and (f) present clustering results of foreground objects derived from DBSCAN.}
\label{fig:1}
\vspace{-0.25in}
\end{figure} 

However, even when considering such temporal information, the lack of proper consideration for instance-level information sometimes leads to points belonging to a single object being categorized into different semantic classes.
Specifically, as illustrated in Fig.~\ref{fig:1}(a), the segmentation results of large vehicles are prone to truncation because the network typically focuses on point-wise classification while ignoring instance-level comprehension. Then, as exhibited in Fig.~\ref{fig:1}(b), even though the motion state of objects is predicted accurately at a certain moment, it is still difficult for a model to ensure the consistency of segmentation in adjacent time.
Thus, how can a model achieve consistent results in both temporal and spatial space? One promising method may be clustering. For outdoor scenes with a sparse distribution of foreground objects, the clustering approaches such as DBSCAN~\cite{DBSCAN} could provide complete object appearance (in Figs.~\ref{fig:1}(e) and (f)), which is suitable for guiding the network in generating segmentation results that satisfy spatio-temporal consistency.

Building upon this idea, we design a dual-branch segmentation network, called 4D-CS, which views historical features as prior knowledge and further develops cluster-based branch to improve the consistency of segmentation through instance information.
Specifically, in the \textit{point-based branch}, we extract point features and adopt a Multi-View Temporal Fusion (MTF) module to enhance them using historical features.  
Unlike~\cite{templidarseg,memoryseg}, which leverage a memory feature that may accumulate noise to transmit historical knowledge, MTF only considers the most recent historical feature to prevent the sustained influence of incorrect information during inference. 
Moreover, instead of utilizing a single view in~\cite{templidarseg}, MTF applies past multi-view observations to supplement the spatial features.
For the \textit{cluster-based branch}, the intent is to generate cluster labels and utilize them to integrate instance information from point-wise features. 
Thus, we first employ voxel-based voting to transfer past semantic predictions to the current frame, then use DBSCAN~\cite{DBSCAN} to group foreground objects from multiple frames and aggregate cluster features with pooling. 
However, the cluster labels do not always fully represent the complete appearance of objects, especially with sparse or occluded point clouds.
To address this, we propose a Temporal Cluster Enhancement (TCE) module to collect cluster features from the past frame, improving the integrity of object information.
Finally, to strengthen the semantic consistency of points within the same object, we present an Adaptive Prediction Fusion (APF) module in the \textit{point-cluster fusion} stage, which adaptively fuses segmentation results from two branches.

The main contributions of 4D-CS are as follows: 
\begin{itemize}
    \item  A dual-branch segmentation network using explicit clustering information to resolve inconsistent predictions of point categories within the same foreground object.

    \item A novel strategy for obtaining cluster labels, accompanied by three modules: the Multi-view Temporal Fusion, Temporal Cluster Enhancement and Adaptive Prediction Fusion,  designed to improve segmentation by utilizing instance information and integrating temporal features. 

    \item The state-of-the-art performance on multi-scan semantic and moving object segmentation on the SemanticKITTI and nuScenes datasets. Our code will be released soon.
\end{itemize} 

\section{Related Work}
\subsection{Single-scan Semantic Segmentation} 

Existing single-scan semantic segmentation algorithms can be classified into point-based, voxel-based, projection-based, and mixture representations.
\textit{Point-based algorithms}~\cite{ptv2,kpconv,pointnet++,PointNet} directly encode features from the raw points.
For example, PointNet~\cite{PointNet} leverages multi-layer perceptrons (MLP) to extract point-wise features. Then, to improve local structure perception, PointNet++~\cite{pointnet++} introduces a hierarchical network for multi-scale information aggregation, while KPConv~\cite{kpconv} utilizes kernel-based point convolution to encode local spatial features. 
However, point-based algorithms are computationally intensive, which constrains their applicability in outdoor scenarios.
In contrast, \textit{voxel-based methods}\cite{cylinder3d,sphereformer} transform unordered points into regular voxels to reduce computational costs. 
Cylinder3D~\cite{cylinder3d} divides space into cylindrical partitions and avoids redundant processing on empty voxels by sparse convolution.
Additionally, SphereFormer~\cite{sphereformer} presents a radial window to improve the segmentation results of distant points.
Yet, these methods are sensitive to voxel size, since big voxel causes information loss while small voxel reduces efficiency.
Besides, some \textit{projection-based methods}\cite{cenet,rethinkingRange,polarnet} project point clouds onto 2D planes for feature extraction.
PolarNet\cite{polarnet} maps points to polar grids to encode features. CENet~\cite{cenet} and RangeFormer~\cite{rethinkingRange} convert LiDAR points to range images. The former adds auxiliary heads for the stronger supervision, while the latter solves many-to-one problem with supervised post-processing.
Nevertheless, the projection operation loses much geometric information, limiting segmentation accuracy.
Moreover, some \textit{mixture-based methods}\cite{spvnas,rpvnet} attempt to combine the benefits of various representations.
RPVNet~\cite{rpvnet} proposes a range-point-voxel fusion network, integrating the features of different representations by weighted calculation.
However, these methods only use a single scan and neglect temporal relationships, resulting in suboptimal segmentation when the LiDAR points are sparse or occluded.

\subsection{Multi-scan Semantic and Moving Object Segmentation}
Since multiple frames contain complete object appearance
and reflect the motion state of objects, some works attempt to leverage the spatio-temporal information from multiple scans to enhance the completeness and continuity of segmentation. To reuse valuable history knowledge, MemorySeg~\cite{memoryseg} maintains a voxel-based memory to improve segmentation results for the current frame. Then, SVQNet~\cite{svqnet} completes invisible geometric information at present by searching historical local features surrounding the current points.
Moreover, built upon distillation learning, 2DPASS~\cite{2dpass} conveys image knowledge to assist with point cloud segmentation. TASeg~\cite{TASeg} chooses specific time steps founded on the classification difficulty of each class to stack multiple scans, while also enhancing point features with temporal image features.
\begin{figure*}[t!]
\vspace{0.05in}
\centering
\includegraphics[width=0.95\linewidth]{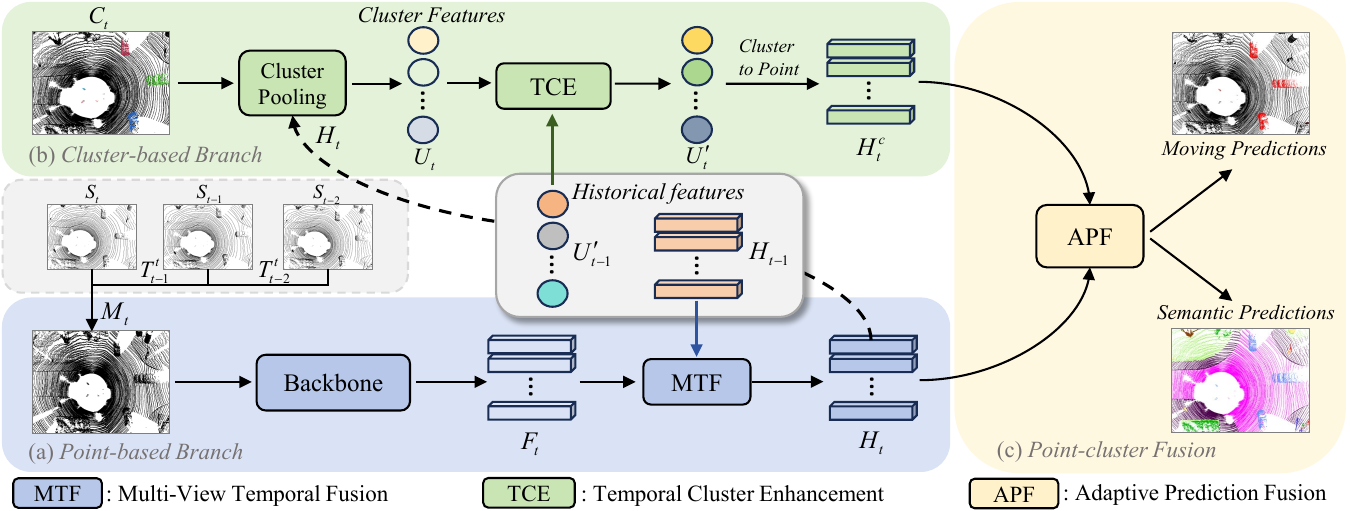}
\vspace{-0.05in}
\caption{The framework of our 4D-CS. (a) In the point-based branch, we extract point-wise features and enhance them using historical knowledge through the MTF module. (b) In the cluster-based branch, cluster labels are first used as additional input to generate initial cluster features. The TCE module then integrates adjacent cluster features across multiple frames to enrich instance information, which is subsequently assigned to the corresponding points. (c) Finally, the segmentation results from the two branches are fused adaptively in Point-cluster Fusion.}
\label{fig:2}
\vspace{-0.15in}
\end{figure*} 

Additionally, some methods have focused on using LiDAR sequences to distinguish the motion state of each point~\cite{4DMOS,Rvmos,MF-MOS}.
4DMOS~\cite{4DMOS} exploits sparse 4D convolution to extract
temporal features and applies a Binary Bayes Filter to merge
prediction results from different time windows. 
RVMOS~\cite{Rvmos} and MF-MOS~\cite{MF-MOS} adopt extra semantic features to assist the model in determining whether the objects are moving or not. However, the above methods do not fully consider semantic
consistency, which may result in points belonging to the same
object having non-uniform categories.

\subsection{Cluster-based Semantic Segmentation}
Currently, several algorithms~\cite{dynaclusterATT,clustering3D} have integrated the cluster concept into segmentation to get better performance. DCTNet~\cite{dynaclusterATT} performs clustering based on feature distance, enabling the network to aggregate the semantically homogeneous points.
Later, \cite{clustering3D} unites clusters with self-supervised learning and processes point features within the same class to generate cluster features. Then, by using contrastive learning, the network could discover latent yet representative subclass patterns.
Unlike the above algorithms that perform clustering at the feature level, we explicitly create the cluster labels of foreground objects and guide the network to yield predictions that are consistent in both spatial and temporal dimensions.

\section{Methodology}
\subsection{Overview} 
In this section, we propose a cluster-assisted method, 4D-CS, which improves the consistency of segmentation results for points belonging to the same object. As displayed in Fig. 2, our method consists of \textit{point-based branch}, \textit{cluster-based branch}, and \textit{point-cluster fusion}.
For the \textit{point-based branch} in Fig. 2(a), we first align multi-scan point clouds using ego-motion and feed them into the backbone network to extract features $F_{t}$.
To leverage past knowledge, we use the Multi-View Temporal Fusion (MTF) module to merge temporal features on multiple views, resulting in an enhanced feature $H_t$.
For the \textit{cluster-based branch} in Fig. 2(b), we produce the cluster labels $C_{t}$ based on historical predictions and exploit them to aggregate initial instance features $U_{t}$ from point features $H_t$. 
Later, a Temporal Cluster Enhancement (TCE) module is proposed to integrate temporal cluster features, which are then allocated to foreground points to create refined instance features $H^c_{t}$.
Finally, for the \textit{point-cluster fusion} in Fig. 2(c), we adopt features from both branches to predict segmentation results, then adaptively optimize the semantic categories and motion states of each point in the Adaptive Prediction Fusion (APF) module.

\subsection{Point-based Branch}
As illustrated in Fig.~\ref{fig:2}, we utilize the pose transformation matrix $T^t_{t-n}$ to convert past scans $S_{t-n} (n\in \{1,...,N\})$ into the coordinate system of current points $S_{t}$. By stacking them, we can get dense point clouds $M_{t}=\left\{p_i\right\}^{K}_{i=1} (p_i\in\mathbb{R}^5)$, where each point $p_i$ consists of 3D coordinates $(x,y,z)$, intensity $r$ and distance $d$ from the origin of the LiDAR sensor frame.
During the point feature extraction, we adopt WaffleIron~\cite{waffle} as our backbone network, which first combines K-Nearest Neighbors (KNN) with MLP to get coarse local features for each point. Thereafter, the points are mapped onto 2D planes of different views to extract features, avoiding the computational burden caused by directly processing numerous point clouds.
Specifically, we project point features along the $z$-axis onto the $x$-$y$ plane and exploit 2D convolutions to extract semantic information. Subsequently, we back-project 2D features into point clouds and map them again along the $y$-axis and $x$-axis onto other planes. Through repeating the above process, we can achieve efficient feature extraction and generate point-wise features $F_{t}\in\mathbb{R}^{N_p \times D}$, where $N_p$ is the number of downsampled points.

\begin{figure}[t]
\centering
\includegraphics[width=8.5cm, height=4.8cm]{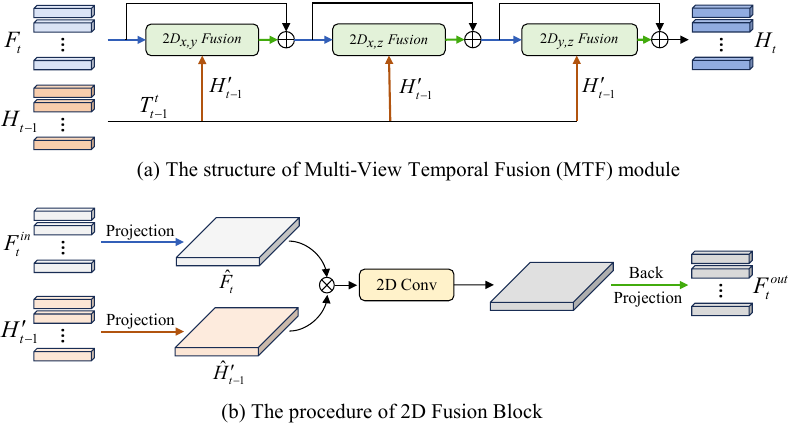}
\vspace{-0.1in}
\caption{In the MTF module shown in (a), we sequentially fuse the current and historical features on the $x$-$y$, $y$-$z$, and $x$-$z$ planes using the 2D Fusion module illustrated in (b) to integrate the 3D spatial features efficiently.}
\label{fig:3} 
\vspace{-0.2in}
\end{figure}

\textit{Multi-View Temporal Fusion:} To leverage temporal information fully, we utilize a MTF module to combine historical information with the current features. Initially, the projection matrix $T^t_{t-1}$ is applied to transform historical features $H_{t-1}$ to the current frame's coordinate system.
Then, as shown in Fig. \ref{fig:3}(a), we sequentially feed the transformed features $H'_{t-1}$ and $F_{t}$ into 2D fusion blocks corresponding to $x$-$y$, $x$-$z$ and $y$-$z$ planes for temporal fusion.
The procedure of 2D fusion is shown in Fig. \ref{fig:3}(b). First, the point feature inputs $(H'_{t-1}, F^{in}_{t})$ are projected into 2D grids along a specific coordinate axis. Later, we average point features within the same grid and get 2D features $(\hat{H}'_{t-1}, \hat{F}_{t})$ of size $H \times W \times D$. Next, they are combined along channel dimension, and a 1$\times$1 convolution is used to perform feature fusion. The 2D features are then back-projected to the corresponding 3D points to replace the original features.
Finally, by carrying out the mentioned step across different views, we could embed historical knowledge and obtain enhanced features $H_{t}\in\mathbb{R}^{N_p \times D}$, thereby reducing information loss due to occlusion.

\subsection{Cluster-based Branch}
Most semantic segmentation networks~\cite{memoryseg,svqnet} typically lack instance-level perception, which will lead to inconsistent semantic predictions for points belonging to the same object (Figs.~\ref{fig:1}(a) and (b)). To address this, we aim to utilize cluster results from multi-scan as additional information to enhance spatio-temporal consistency in semantic segmentation.

\textit{Cluster Label Generation:} Due to the continuity of point cloud sequences, we can adopt ego-motion to align past scans with the current points and assign historical predictions to the present frame. Then, for points categorized as foreground, we can employ DBSCAN to segment them into multiple clusters and acquire cluster labels (Figs.~\ref{fig:1}(e) and (f)).

\begin{figure}[t]
\vspace{0.05in}
\centering
\includegraphics[width=8.2cm, height=8.4cm]{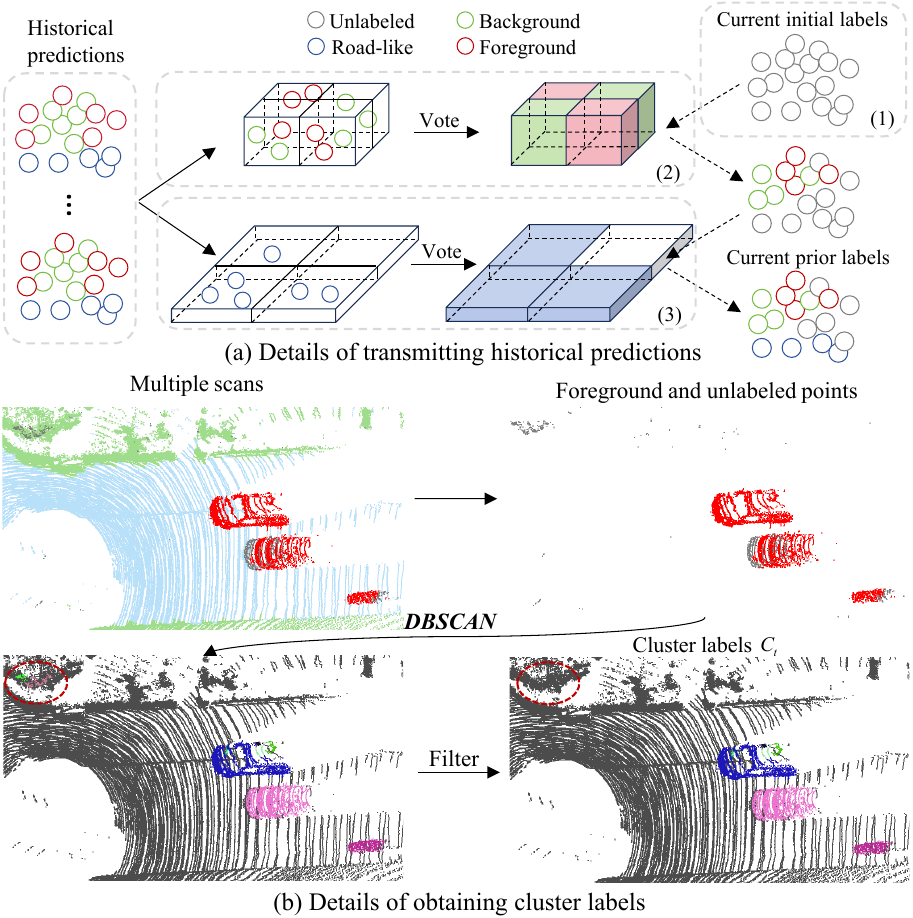}
\vspace{-0.1in}
\caption{The illustration of cluster label generation. We first leverage voxels to transfer historical semantic predictions to the current points, and then the DBSCAN is used to generate the clusters of foreground objects.}
\label{fig:4}
\vspace{-0.2in}
\end{figure}

Specifically, as shown in Fig.~\ref{fig:4}(a), we transfer historical semantic predictions to the current points by the following steps:
(1) \textit{Label Initialization:} Due to focusing on the consistency of foreground segmentation, we map historical predictions to background, foreground, and road-like. Meanwhile, all points in the $t$ frame are initialized as ``unlabeled".
(2) \textit{Non-ground Label Assignment:} At first, we transfer historical non-ground points to the coordinate system of $t$ frame by a transformation matrix. Next, we separate the 3D space into multiple voxels of size $(w, l, h)$ and feed historical points into corresponding voxels. Through the max-voting operation, the voxel class is assigned based on the most frequent category among its points. Thereafter, we allocate the voxel classes to the current frame based on the coordinate relationships.
(3) \textit{Ground Label Assignment:}
If translation occurs between two frames, the ground points in the current frame may not have nearby corresponding points from historical frames, resulting in many ground points remaining unlabeled in small voxels of step (2). Thus, we use larger and flatter voxels $(w', l', h')$ to assign road-like labels to ``unlabeled" points.

To obtain foreground cluster results that encompass spatio-temporal information, we cluster dense points $M_{t}$ which is the stacked point clouds of multiple scans as shown in Fig. \ref{fig:4}(b).
Yet, since the foreground objects may be moving, some dynamic points are still unlabeled after label assignment. Therefore, we retain the points categorized as the foreground and unlabeled across multiple scans, and then process them using DBSCAN to obtain initial cluster results $\widetilde{C}_{t}=\{c_{i}\}^{N_c}_{i=1}$, where $N_c$ means the number of clusters. 
Except for dynamic objects, new background points observed in the current scan may also be classified as unlabeled, leading to some clusters belonging to the background.
To tackle this, we loop through the classes of all points within each cluster and retain clusters that contain foreground points. Then, the filtered clusters are denoted as $C_{t}=\{c_{i} \mid \exists p_{i}^{j} \in c_{i} \text{ and } L(p_{i}^{j})=\text{“foreground”}\}$, where $p_{i}^{j}$ is the $j$-th point in $c_{i}$, and $L$ represents predicted category of each point.

\textit{Instance Feature Aggregation:} This part aims to gather the features $H_t$ of point-based branch based on cluster labels $C_{t}$ to acquire instance information. A simple but effective way is to average all point features within the same cluster to yield the cluster features $U_t=\left\{u_{i}\in\mathbb{R}^D\right\}^{N_c}_{i=1}$. Meanwhile, the 3D coordinates of points are also averaged to produce the cluster centers $G_t=\left\{g_{i}\in\mathbb{R}^3\right\}^{N_c}_{i=1}$.
However, due to the sparsity or occlusion of point clouds, DBSCAN may separate points of the same object into multiple clusters, resulting in the cluster feature $u_{i}$ not reflecting the instance information well. Thus, we propose a Temporal Cluster Enhancement (TCE) module to supplement cluster features with neighbors across multi-frame and improve the integrity of cluster information.

In TCE, we project historical cluster centers $G_{t-1}$ to the current coordinate system by transformation matrix $T_{t-1}^{t}$ and combine it with the current cluster to acquire the new cluster centers $G_{t-1}^{t}=\left\{g'_{j}\right\}^{M}_{j=1}$ and corresponding features $U_{t-1}^{t}=\left\{u'_{j}\right\}^{M}_{j=1}$, where $M$ represents the total number of clusters across multiple frames.
Then, we leverage KNN to search for neighbors $\mathcal{N}(g_{i})=\left\{(g'_{j},u'_{j}) \mid g'_{j} \in Neighborhood(g_{i})\right\}$ in $G_{t-1}^{t}$, which are adjacent to the current cluster centers $G_t$.
For the given cluster $c_{i}=(g_{i},u_{i})$, we utilize a linear layer to map feature $u_{i}$ into \textit{query} $q_{i}$. And the feature $u'_{j}$ of the neighboring clusters $c'_{j}=(g'_{j},u'_{j}) \in \mathcal{N}(g_i)$ is projected into \textit{key} $k_{j}$ and \textit{value} $v_{j}$ vectors. After that, we divide the channel of $v_{j}\in\mathbb{R}^{D}$ into $h$ groups ($1 \le h \le D$) and employ Grouped Vector Attention~\cite{ptv2} to aggregate cluster features $u'_{j}$ near $c_{i}$, which is denoted as:
\begin{equation}
w_{ij} = \omega(k_j-q_i+\delta_{bias}(g'_j-g_i)),
\end{equation}
\begin{equation}
u^{attn}_{i}=\sum^{\mathcal{N}(g_{i})}_{c'_{j}}\sum^{h}_{l=1}\sum^{D/h}_{m=1}\text{Softmax}(W_{i})_{jl}v^{lD/h+m}_{j},
\end{equation}
where $\delta_{bias}$ is the positional encoding function, and $W_{i}$ is the collection of all $w_{ij}$ for different neighbors $c'_{j}$ of cluster $c_{i}$. $\omega:\mathbb{R}^{D} \rightarrow \mathbb{R}^{h}$ means the learnable grouped weight encoding.
Meanwhile, enhanced cluster features are formulated as $U'_{t}=\left\{u^{attn}_{i}\right\}^{N_c}_{i=1}$.
In the end, we distribute the cluster features $U'_{t}\in\mathbb{R}^{N_c \times D}$ to the corresponding foreground points. For residual points, we fill zero as their features and get a final point-wise cluster feature $H^{c}_{t}\in\mathbb{R}^{N_p \times D}$ that has the same size as $H_{t}$.

\subsection{Point-cluster Fusion}
\begin{figure}[t]
\vspace{0.08in}
\centering
\includegraphics[width=8.2cm, height=3.3cm]{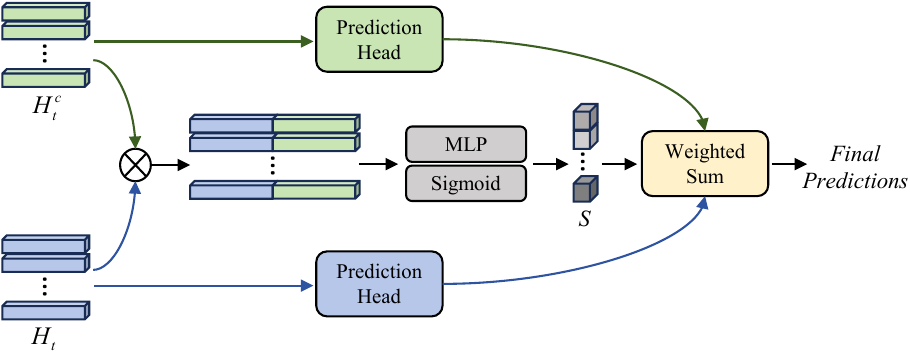}
\vspace{-0.1in}
\caption{Illustration of Adaptive Prediction Fusion (APF) module. We adopt different heads to estimate logits for point features from different branches while combining these two features to compute confidence scores. Then, we perform a weighted sum of logits to generate the final prediction results.}
\label{fig:5}
\vspace{-0.1in}
\end{figure}

To combine the semantic features and instance information of two branches and get spatio-temporal consistent segmentation results, we propose an Adaptive Prediction Fusion (APF) module to adaptively merge the predictions of two branches in the point-cluster fusion stage.
As illustrated in Fig.~\ref{fig:5}, for the features $(H_t, H^c_t)$ from different branches, we adopt specific heads to estimate semantic categories and motion states for each point, obtaining the semantic logits $(P_{sem}, P^{c}_{sem})$ and motion logits $(P_{mov}, P^{c}_{mov})$.
Later, to weight the predicted logits from two branches, we join the point features $(H_t, H^c_t)$ along the channel dimension and compute confidence scores $S\in \{S_{sem}, S_{mov}\}$, with values ranging from 0 to 1, through two MLPs that do not share weights.
\begin{equation}
S=\text{Sigmoid}(\text{MLP}(\text{Concat}(H_t, H^c_t))).
\end{equation}
Afterward, the confidence scores $(S_{sem}, S_{mov})$ are employed to merge predicted logits of two branches adaptively, which can be represented by the following formula:
\begin{equation}
P^{final}_{sem}=(1-S_{sem})\cdot P_{sem}+S_{sem} \cdot P^{c}_{sem},
\end{equation}
\begin{equation}
P^{final}_{mov}=(1-S_{mov})\cdot P_{mov}+S_{mov} \cdot P^{c}_{mov}.
\end{equation}

\subsection{Loss Function}
During the training process, given the ground truth labels, we adopt the predicted semantic logits $P^{final}_{sem}$ and the motion logits $P^{final}_{mov}$ of each point to calculate the losses as follows: 
\begin{equation}
\mathcal{L}=\mathcal{L}^{sem}_{ce}+\mathcal{L}^{sem}_{ls}+\mathcal{L}^{mov}_{ce}+\mathcal{L}^{mov}_{ls},
\end{equation}
where $\mathcal{L}^{sem}_{ce}$ and $\mathcal{L}^{mov}_{ce}$ are cross-entropy losses for semantic and motion prediction, respectively.
$\mathcal{L}^{sem}_{ls}$ and $\mathcal{L}^{mov}_{ls}$ are the Lovasz Softmax Loss~\cite{lovasz} for semantic and motion results.
This loss function serves as a differentiable surrogate, aiming to optimize the Intersection over Union (IoU) that is used to measure segmentation quality, thereby compensating for the shortcomings of cross-entropy loss in the optimization objective.

\begin{table*}[t]
\vspace{0.05in}
\scriptsize
\renewcommand\tabcolsep{1.3pt}
\caption{The results on the multi-scan semantic segmentation of the SemanticKITTI test set. (m) indicates moving. The highest IoU for each category is bolded. $\Delta$ means comparison with baseline.}
\vspace{-0.05in}
~\label{tab:SemanticKITTI}
\centering
\begin{tabular}{@{}c|c|ccccccccccccccccccccccccc@{}}
\thickhline
\rule{0pt}{42pt}
\textbf{Methods}      & \rotatebox{90}{mIoU(\%)}      & \rotatebox{90}{car}           & \rotatebox{90}{bicycle}       & \rotatebox{90}{motorcycle}    & \rotatebox{90}{truck}         & \rotatebox{90}{other-vehicle} & \rotatebox{90}{person}        & \rotatebox{90}{bicyclist}    & \rotatebox{90}{motorcyclist}  & \rotatebox{90}{road}          & \rotatebox{90}{parking}       & \rotatebox{90}{sidewalk}      & \rotatebox{90}{other-ground}  & \rotatebox{90}{building}      & \rotatebox{90}{fence}         & \rotatebox{90}{vegetation}    & \rotatebox{90}{trunk}         & \rotatebox{90}{terrain}       & \rotatebox{90}{pole}          & \rotatebox{90}{traffic sign}  & \rotatebox{90}{car (m)}       & \rotatebox{90}{bicyclist (m)} & \rotatebox{90}{person (m)}    & \rotatebox{90}{motorcyc. (m)} & \rotatebox{90}{other-veh. (m)} & \rotatebox{90}{truck (m)}     \\ \hline \hline
TemporalLidarSeg~\cite{templidarseg}      & 47.0          & 92.1          & 47.7          & 40.9          & 39.2          & 35.0          & 14.4          & 0.0          & 0.0           & 91.8          & 59.6          & 75.8          & 23.2          & 89.8          & 63.8          & 82.3          & 62.5          & 64.7          & 52.6          & 60.4          & 68.2          & 42.8          & 40.4          & 12.9             & 12.4              & 2.1           \\
TemporalLatticeNet~\cite{TemporalLatticeNet}   & 47.1          & 91.6          & 35.4          & 36.1          & 26.9          & 23.0          & 9.4           & 0.0          & 0.0           & 91.5          & 59.3          & 75.3          & 27.5          & 89.6          & 65.3          & 84.6          & 66.7          & 70.4          & 57.2          & 60.4          & 59.7          & 41.7          & 51.0          & 48.8             & 5.9               & 0.0           \\
Meta-RangeSeg~\cite{metarangeseg}        & 49.7          & 90.8          & 50.0          & 49.5          & 29.5          & 34.8          & 16.6          & 0.0          & 0.0           & 90.8          & 62.9          & 74.8          & 26.5          & 89.8          & 62.1          & 82.8          & 65.7          & 66.5          & 56.2          & 64.5          & 69.0          & 60.4          & 57.9          & 22.0             & 16.6              & 2.6           \\
KPConv~\cite{kpconv}               & 51.2          & 93.7          & 44.9          & 47.2          & 42.5          & 38.6          & 21.6          & 0.0          & 0.0           & 86.5          & 58.4          & 70.5          & 26.7          & 90.8          & 64.5          & 84.6          & 70.3          & 66.0          & 57.0          & 53.9          & 69.4          & 67.4          & 67.5          & 47.2             & 4.7               & 5.8           \\
Cylinder3D~\cite{cylinder3d}            & 52.5          & 94.6          & 67.6          & 63.8          & 41.3          & 38.8          & 12.5          & 1.7          & 0.2           & 90.7          & 65.0          & 74.5          & 32.3          & 92.6          & 66.0          & 85.8          & 72.0          & 68.9          & 63.1          & 61.4          & 74.9          & 68.3          & 65.7          & 11.9             & 0.1               & 0.0           \\
MarS3D~\cite{mars3d}              & 52.7          & 95.1          & 49.2          & 49.5          & 39.7          & 36.6          & 16.2          & 1.2          & 0.0           & 89.9          & 66.8          & 74.3          & 26.4          & 92.1          & 68.2          & 86.0          & 72.1          & 70.5          & 62.8          & 64.8          & 78.4          & 67.3          & 58.0          & 36.3             & 10.0              & 5.1           \\
Cluster3DSeg~\cite{clustering3D}          & 54.7          & 95.3          & 55.9 & 52.9 & 42.7          & 38.7          & 15.5          & 0.0          & 3.0           & 91.4          & 66.1          & 76.9          & 27.8          & 91.4          & 66.1          & 86.5          & 72.7          & 71.6          & 64.0          & 68.0          & 81.7          & 68.2 & 61.8          & 46.0             & 11.2              & 42.7          \\
MemorySEG~\cite{memoryseg}           & 58.3          & 94.0          & 68.3 & \textbf{68.8} & 51.3          & 40.9          & 27.0          & 0.3          & 2.8           & 89.9          & 64.3          & 74.8          & 29.2          & 92.2          & 69.3          & 84.8          & 75.1          & 70.1          & 65.5          & 68.5          & 71.7          & 74.4 & 71.7          & 73.9             & 15.1              & 13.6          \\
SVQNet~\cite{svqnet}               & 60.5          & 96.1          & 64.4          & 60.3          & 40.4          & \textbf{60.9} & 27.4          & 0.0          & 0.0           & \textbf{93.2} & \textbf{71.6} & \textbf{80.5} & 37.0 & \textbf{93.7} & 72.6 & \textbf{87.3} & \textbf{76.7} & 72.3          & \textbf{68.4} & 71.0          & 80.5          & 72.4          & \textbf{84.7} & \textbf{91.0}    & 7.5               & 3.9           \\ 
2DPASS~\cite{2dpass} & 62.4          & 96.2          & 63.6          & 63.7          & 48.2          & 52.7 & \textbf{35.4}          & \textbf{7.9}          & \textbf{62.0}           & 89.7 & 67.4 & 74.7 & \textbf{40.0} & 93.6 & \textbf{72.9} & 86.2 & 73.9 & 71.0          & 65.0 & 70.5          & 82.1          & 71.2          & 80.3 & 73.1    & 3.8               & 16.1           \\ 
\hline 
\rule{0pt}{7pt}
WaffleIron~\cite{waffle}              & 58.4          & 96.0          & \textbf{69.0}          & 66.8          & 39.9          & 42.3          & 33.4          & 0.4          & 0.0           & 90.6          & 67.9          & 75.3          & 27.7          & 93.3          & 70.6          & 86.6          & 73.6          & 71.9          & 63.9          & 69.0          & 84.2          & \textbf{76.5}          & 70.8          & 45.5            & 20.8              & 24.7 \\
\textbf{4D-CS (Ours)} & \textbf{63.7} & \textbf{96.7} & 66.0          & 66.5          & \textbf{62.4} & 59.3          & 33.7 & 6.7 & 15.0 & 90.4          & 68.3          & 75.3          & 32.6          & 93.4          & 71.3          & 87.0          & 73.9          & \textbf{72.5} & 65.3          & \textbf{71.3} & \textbf{86.0} & 72.3          & 76.6          & 64.6             & \textbf{35.5}     & \textbf{50.9} \\ 
Improvements $\Delta$ & 
\textcolor{darkgreen}{+5.3} & \textcolor{darkgreen}{+0.7} & -3.0 & -0.3 & \textcolor{darkgreen}{+22.5} & \textcolor{darkgreen}{+17.0} & \textcolor{darkgreen}{+0.3} & \textcolor{darkgreen}{+6.3} & \textcolor{darkgreen}{+15.0} & -0.2 & \textcolor{darkgreen}{+0.4} & +0.0 & \textcolor{darkgreen}{+4.9} & \textcolor{darkgreen}{+0.1} & \textcolor{darkgreen}{+0.7} & \textcolor{darkgreen}{+0.4} & \textcolor{darkgreen}{+0.3} & \textcolor{darkgreen}{+0.6} & \textcolor{darkgreen}{+1.4} & \textcolor{darkgreen}{+2.3} & \textcolor{darkgreen}{+1.8} & -4.2 & \textcolor{darkgreen}{+5.8} & \textcolor{darkgreen}{+19.1} & \textcolor{darkgreen}{+14.7} & \textcolor{darkgreen}{+26.2} \\ 
\thickhline
\end{tabular}
\end{table*}

\begin{table*}[t]
\scriptsize
\renewcommand\tabcolsep{2.8pt}
\caption{The results of multi-scan semantic segmentation on nuScenes validation set. (m) indicates moving.}
\vspace{-0.05in}
~\label{tab:Nusc_semantic}
\centering
\begin{tabular}{@{}c|c|cccccccccccccccccccccccc@{}}
\thickhline
\rule{0pt}{38pt}
\textbf{Methods}      & \rotatebox{90}{mIoU(\%)}      & \rotatebox{90}{barrier}           & \rotatebox{90}{bicycle}       & \rotatebox{90}{bus}    & \rotatebox{90}{car}         & \rotatebox{90}{construction} & \rotatebox{90}{motorcycle}        & \rotatebox{90}{pedestrian}    & \rotatebox{90}{traffic cone}  & \rotatebox{90}{trailer}          & \rotatebox{90}{truck}       & \rotatebox{90}{driveable}      & \rotatebox{90}{other flat}  & \rotatebox{90}{sidewalk}      & \rotatebox{90}{terrain}         & \rotatebox{90}{manmade}    & \rotatebox{90}{vegetation}         & \rotatebox{90}{car (m)}       & \rotatebox{90}{bus (m)}          & \rotatebox{90}{truck (m)}  & \rotatebox{90}{const. (m)}       & \rotatebox{90}{trailer (m)} & \rotatebox{90}{motor. (m)}    & \rotatebox{90}{bicyc. (m)} & \rotatebox{90}{person (m)}\\ \hline \hline
MarS3D~\cite{mars3d}     & 54.3      & 70.5          & 24.7          & 60.0           & \textbf{79.9}          & 32.0          & 34.9          & \textbf{51.3}           & 53.0          & 10.4           & 66.0          & 95.4          & 59.9          & 72.7          & \textbf{75.8}          & 87.2          & \textbf{86.1}          & 66.5          & 48.0          & 52.4          & 0.0          & 23.1          & 69.0           & 9.7           & 72.7                           \\
SegNet4D~\cite{segnet4d}    & 57.9     & 77.4          & 32.6          & 63.8          & 73.8          & 41.1          & 44.0          & 51.2           & 63.2          & 42.3           & 74.2          & 96.2          & 69.3          & 74.2          & 73.5          & 64.6          & 55.9          & 68.6          & 51.3          & 59.4          & 0.0         & 27.2          & 74.3         & 40.8          & 72.4                      \\
WaffleIron~\cite{waffle}    & 65.7     & 78.5          & 48.3          & 69.8          & 72.8          & 50.6         & 59.1          & 49.2        & 69.9          & 54.3           & 56.0          & \textbf{96.9}          & 73.6          & \textbf{75.5}          & 74.0         & 87.9          & 85.4          & 70.9         & 62.2          & 49.4         & \textbf{0.3}         & 59.5          & 89.7         & \textbf{70.2}         & \textbf{73.5}                      \\
\hline 
\rule{0pt}{6pt}
\textbf{4D-CS (Ours)}      & \textbf{67.3}         & \textbf{78.8}          & \textbf{51.7}          & \textbf{77.5}          & 78.9          & \textbf{51.6}          & \textbf{61.1}          & 40.4          & \textbf{70.4}           & \textbf{56.5}          & \textbf{75.4}          & 96.8          & \textbf{73.7}          & 75.0         & 73.7          & \textbf{88.0}          & 85.4          & \textbf{75.2}          & \textbf{65.7}          & \textbf{64.0}          & 0.2          & \textbf{61.3}          & \textbf{90.4}          & 55.2             & 69.1                      \\
\thickhline
\end{tabular}
\vspace{-0.1in}
\end{table*}

\section{Experiment}
\subsection{Dataset}
SemanticKITTI~\cite{semantickitti} is a widely used dataset for semantic understanding in outdoor scenes. It utilizes 64-beam LiDAR to collect point clouds and consists of 22 LiDAR sequences, with sequences 00 to 10 as the training set (sequence 08 as the validation set) and sequences 11 to 21 as the testing set. The semantic segmentation task is separated into single-scan (19 categories) only distinguishing object classes and multi-scan (25 categories) extra required to identify motion states of the foreground objects. 
Besides, the SemanticKITTI-MOS is another benchmark that only determines the dynamic and static states of points.
Moreover, nuScenes~\cite{nuScenes} is composed of 1,000 driving scenes collected by a 32-beam LiDAR sensor and provides 16 semantic classes. Then, following methods\cite{mars3d,segnet4d}, we use ground-truth 3D bounding boxes to create 8 moving categories additionally.

\subsection{Evaluation Metric} 
We adopt the intersection over the union (IoU) to evaluate different methods. The IoU is defined as $\frac{TP}{TP+FP+FN}$, where the $TP$, $FP$, and $FN$ denote the true positive, false positive, and false negative of predictions.
For the multi-scan benchmark, we adopt mIoU as the evaluation metric, which denotes the IoU of all classes. For the MOS benchmark, we apply the IoU of the moving objects as the evaluation metric.

\subsection{Implementation Details} 
During the training and testing process, we use three consecutive frames of point clouds as input for SemanticKITTI dataset. 
For nuScenes dataset, in which the LiDAR operates at 20 Hz, we choose three frames with a temporal stride of 2 to better capture object motion. We adopt WaffleIron~\cite{waffle} with $L=48$ layers as a backbone network.
Similar to~\cite{waffle}, we downsample point clouds by keeping only one point per voxel of size 10 $cm$. For the hyperparameters of WaffleIron, we utilize $D=256$ and a grid resolution $\rho$ of 40 $cm$ for SemanticKITTI, and $D=384$ with 60 $cm$ grid for nuScenes.
For cluster label generation, the voxel size of non-ground label assignment is set to $(0.2m,0.2m,0.2m)$, while the voxel size of ground label assignment is $(10.0m,10.0m,0.2m)$.
Moreover, we train the network without historical features for 45 epochs using two NVIDIA RTX 4090 GPUs. Afterward, the backbone is frozen, and the residual modules are trained for an additional 45 epochs. The AdamW~\cite{adamw} is adopted to optimize the network with a weight decay of 0.003 and batch size of 6.
Besides, our data augmentation strategy includes random flipping, rotation, scaling and instance cutmix with polarmix~\cite{waffle}.

\begin{table}[t]
\renewcommand\tabcolsep{4pt}
\caption{ Performance comparison on the validation and test set of SemanticKITTI-MOS. \dag \ indicates training on both SemanticKITTI and KITTI-road datasets, while * denotes methods using semantic labels.}
~\label{tab:SemanticKITTI-MOV}
\centering
\begin{tabular}{c|c|c}
\thickhline
\rule{0pt}{8pt}
Methods  & IoU$_{M}$ (Validation 08) & IoU$_{M}$ (Test 11-21) \\ \hline \hline
\rule{0pt}{8pt}
MotionSeg3D~\cite{MotionSeg3D}  & 71.4 & 64.9 \\
4DMOS~\cite{4DMOS}    & 77.2 & 65.2 \\ 
LMNet~\cite{LMNet} & 67.1 & 62.5 \\
RVMOS*~\cite{Rvmos}  & 71.2 & 73.3 \\ 
InsMOS*~\cite{insmos}  & 73.2 & 70.6 \\ 
InsMOS*\dag~\cite{insmos}  & 69.4 & 75.6 \\ 
MotionBEV~\cite{MotionBEV}  & 76.5 & {69.7} \\
MF-MOS*~\cite{MF-MOS}  & 76.1 & 76.7\\  
\hline 
\rule{0pt}{8pt}
\textbf{4D-CS*(Ours)}  & \textbf{80.9} & \textbf{83.5}\\
\thickhline
\end{tabular}
\vspace{-0.15in}
\end{table}

\subsection{Evaluation Results}
\textit{Quantitative Results:} As displayed in Tab.~\ref{tab:SemanticKITTI} and Tab.~\ref{tab:Nusc_semantic}, we compare our algorithm with other methods on the multi-scan semantic segmentation of SemanticKITTI and nuScenes. The results demonstrate that the proposed 4D-CS achieves state-of-the-art performance in terms of mIoU. Compared to the baseline~\cite{waffle}, we achieve a significant improvement for large foreground objects with a 22.5\% increase for trucks, 17.0\% for other vehicles, 26.2\% for dynamic trucks, and 14.7\% for moving other vehicles. 
In Tab.~\ref{tab:Nusc_semantic}, our method also achieves IoU improvement of most foreground objects, especially for large ones, such as trucks and buses.
This demonstrates that explicit clustering priors can help the network to focus on the complete spatial information of objects, rather than relying on the local features obtained from a limited receptive field as other algorithms do, resulting in better foreground object segmentation results.
Moreover, we compare the performance of our approach on the MOS benchmark of the SemanticKITTI in Tab.~\ref{tab:SemanticKITTI-MOV}. 
Our approach surpasses the state-of-the-art work MF-MOS~\cite{MF-MOS} by 6.8\% IoU$_{M}$ on the testing set. This validates that transmitting historical features at both point and instance levels can not only improve the integrity of segmentation but also enhance the model's ability to identify object motion states.

\begin{figure}[t!]
\vspace{0.05in}
\centering
\includegraphics[width=8.4cm, height=5.0cm]{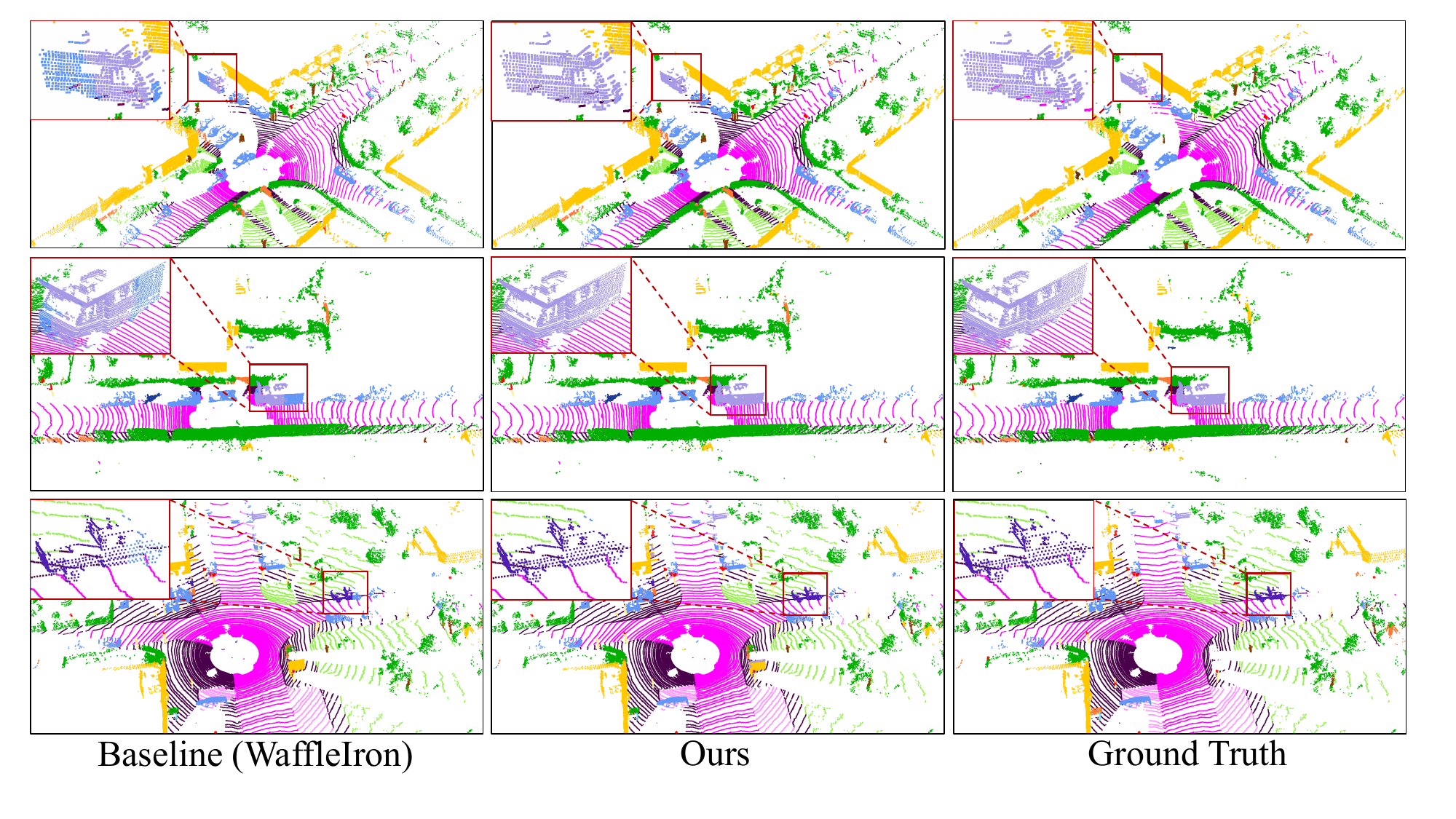}
\vspace{-0.1in}
\caption{The visualization of semantic segmentation results on the validation set of SemanticKITTI. 
We highlight the area that reflects the advantages of our method, which is displayed in the upper left corner.}
\label{fig:6}
\vspace{-0.05in}
\end{figure}

\begin{figure}[t]
\centering
\includegraphics[width=8.4cm, height=5.0cm]{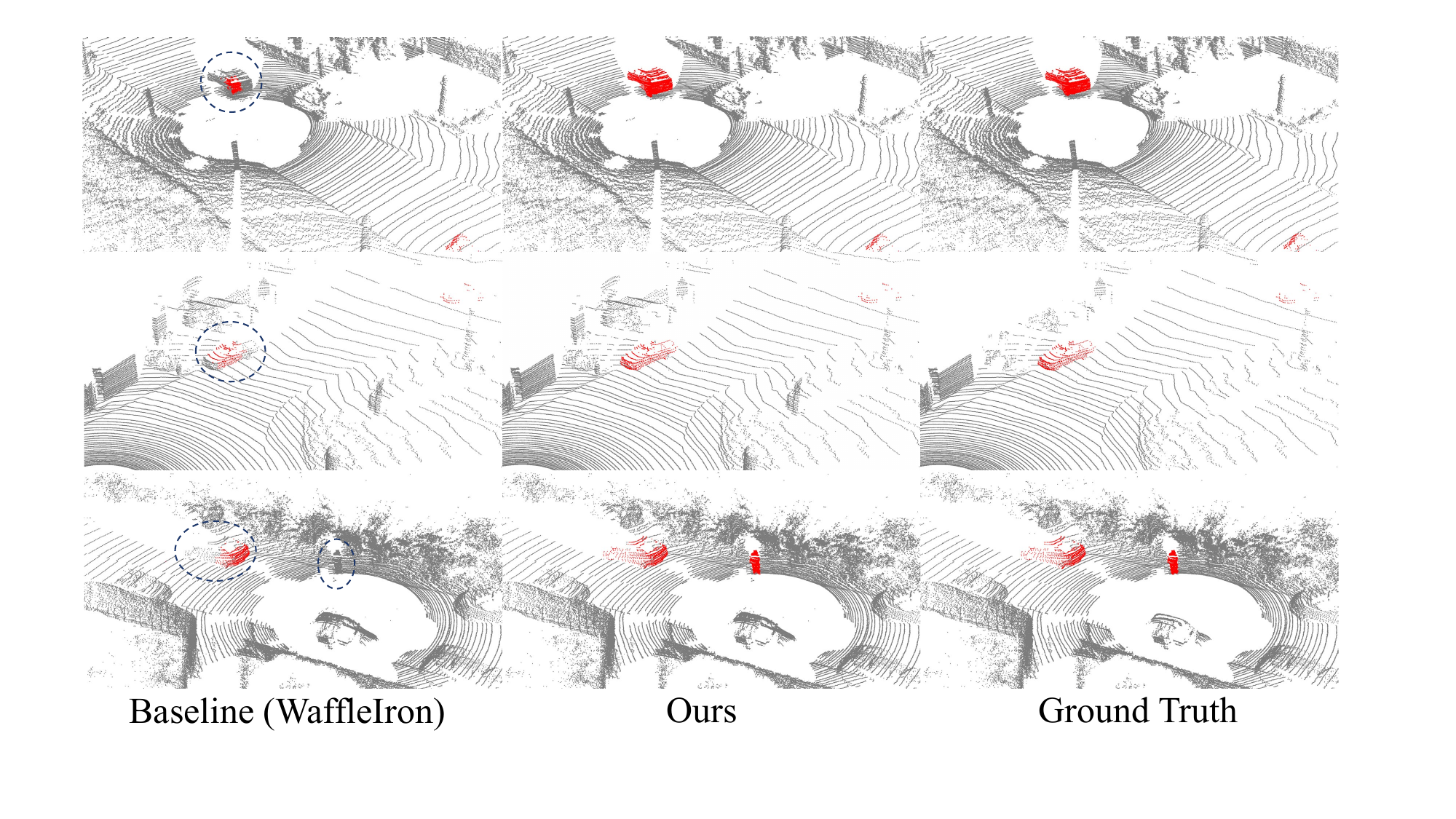}
\vspace{-0.1in}
\caption{The visualization of moving object segmentation on the validation set of SemanticKITTI. We mark the poor predictions of baseline with blue dashed circles.}
\label{fig:7}
\vspace{-0.2in}
\end{figure}

\textit{Qualitative Comparisons:} Semantic qualitative results are illustrated in Fig.~\ref{fig:6}. It shows that the segmentation results of the baseline network for large objects are prone to truncation due to lacking the ability for instance perception. In contrast, our method can achieve consistent segmentation results after introducing cluster information. Additionally, for the moving qualitative results displayed in Fig.~\ref{fig:7}, the baseline model still struggles to segment moving objects completely, whereas our method successfully achieves this. Overall, our approach has a stronger capability to accurately and consistently recognize the categories and motion states of foreground objects.

\subsection{Ablation Studies}
In this section, we conduct comprehensive ablation experiments on the validation set of the SemanticKITTI dataset. 

\textit{Model Components:} As shown in Tab.~\ref{tab:Component Ablation}, we first employ backbone~\cite{waffle} to extract features and directly use the output features to generate predictions, resulting in a baseline multi-scan semantic mIoU of 56.4\% and 76.7\% IoU of the moving objects. 
After that, we introduce cluster labels to gather point features and adopt Adaptive Prediction Fusion (APF) module to optimize the prediction results, leading to the improvement of 0.7\% mIoU and 2.5\% IoU$_{M}$. It proves the proposed APF can effectively utilize cluster knowledge to improve instance-level perception and enhance the accuracy of semantic predictions.
Since only foreground objects have motion states, this instance-level perception of the foreground significantly improves the IoU of moving objects.
Later, we exploit TCE* (i.e., Temporal Cluster Enhancement (TCE) module without historical cluster features), leading to better performance and proving the effectiveness of merging nearby cluster features.
In the end, we supply historical features into our pipeline. On the one hand, we adopt Multi-View Temporal Fusion (MTF) module to combine past point features, increasing both mIoU and the IoU$_{M}$ by 0.4\%.
On the other hand, we adopt the TCE module with historical cluster features and improve mIoU by 0.3\% and IoU$_{M}$ by 0.9\%.
These results indicate that fusing historical priors can effectively enhance the current point and cluster features, resulting in better segmentation results.

\textit{Dual-branch Fusion:} As shown in Tab.~\ref{tab:Detailed Ablation}, we compare different strategies for merging information from two branches. It is worth noting that our network contains a TCE* module and a dual-branch fusion module with different strategies in this part. 
First, we directly overwrite point features with cluster features, leading to a performance drop compared to using a weighted sum. 
This is because the foreground point segmentation depends on cluster features, where an error will lead to all points of a cluster being predicted incorrectly.
Then, we concatenate the features from different branches and use MLP for fusion, but the results are inferior to the weighted sum.
Besides, compared to the adaptive weighted fusion of predictions in APF, we attempt a hard fusion of prediction logits from different branches, but it also leads to an unideal result.
This proves that our proposed weight-based soft fusion can avoid damage to the original results caused by some poor cluster features and achieve better predictions.

\begin{table}[t!]
\vspace{0.05in}
\centering
\renewcommand\tabcolsep{4.5pt}
\caption{The effect of different modules on SemanticKITTI validation set. mIoU is the mean IoU of all classes on the semantic segmentation. IoU$_{M}$ is the IoU of moving objects.}
\vspace{-0.06in}
~\label{tab:Component Ablation}
\begin{tabular}{ccccc|c|c}
\thickhline
\rule{0pt}{8pt}
Baseline & APF & TCE* & MTF & TCE & mIoU (\%) & IoU$_{M}$ (\%) \\ \hline \hline
\rule{0pt}{8pt}
\checkmark & & & & & 56.4 & 76.7 \\
\checkmark & \checkmark & & & & 57.1 & 79.2 \\
\checkmark & \checkmark & \checkmark & & & 57.3 & 79.6  \\
\checkmark & \checkmark & \checkmark & \checkmark & & 57.7 & 80.0  \\
\checkmark & \checkmark & & \checkmark & \checkmark & \textbf{58.0} & \textbf{80.9}  \\
\thickhline
\end{tabular}
\end{table}

\begin{table}[t!]
\renewcommand\tabcolsep{9.5pt}
\caption{More detailed ablation experiments on the modules of 4D-CS on SemanticKITTI validation set.}
\vspace{-0.06in}
~\label{tab:Detailed Ablation}
\centering
\begin{tabular}{c|c|c|c}
\thickhline
\rule{0pt}{8pt}
Module  & Strategy & mIoU (\%)  & IoU$_{M}$ (\%)  \\ \hline \hline
\rule{0pt}{8pt}
\multirow{4}{*}{APF} 
                                & Direct Overwrite & 55.6 & 77.4 \\
                                & Feature Fusion & 56.8 & 79.3 \\
                                & Unweighted Sum & 57.1 & 78.7 \\
                                & \textbf{Weighted Sum} & \textbf{57.3} & \textbf{79.6} \\
                                 \hline 
                                 \rule{0pt}{8pt}
\multirow{3}{*}{MTF} & w/o MTF & 57.3 & 79.6 \\
                                 & Only BEV View  & 57.6 & 79.7 \\
                                 & \textbf{Multiple Views} & \textbf{57.7} & \textbf{80.0}  \\
\thickhline
\end{tabular}
\vspace{-0.2in}
\end{table}

\textit{Multi-view Temporal Feature Fusion:} In this section, our network is comprised of the TCE* and APF modules, while we compare historical feature fusion on the single-view and multi-view.
In Tab.~\ref{tab:Detailed Ablation}, compared to the situation without temporal fusion, the method that incorporates historical features shows improvements in both multi-scan semantic mIoU and IoU$_{M}$ of moving objects.
In particular, our proposed multi-view fusion strategy achieves the best performance improvement, proving that multi-view fusion could more effectively reduce information loss and convey temporal cues more completely.

\subsection{Runtime and Memory}
In this section, we employ an NVIDIA RTX 4090 GPU to measure the inference time for multi-scan semantic segmentation on the SemanticKITTI dataset. With three point cloud frames, our baseline method (WaffleIron) takes 117 ms and occupies 8.2 GB of memory. In comparison, our proposed algorithm requires 151 ms for network processing and 5 ms for cluster label generation, utilizing 9.9 GB of memory.

\section{Conclusions}
In this paper, we analyze the limitations of existing multi-scan segmentation methods and propose a novel dual-branch structure, which aims to use cluster information to improve spatio-temporal consistency of segmentation results. 
We first fuse temporal point features by the multi-view representation. Then, we utilize cluster labels to integrate point features and acquire instance information, which is refined by combining neighboring clusters across multiple frames.
Finally, we fuse information from two branches adaptively to optimize the class prediction of each point, thereby boosting the consistency of segmentation. The experiments show that our 4D-CS exceeds the previous state-of-the-art multi-scan semantic and moving object segmentation methods.

\bibliographystyle{ieeetr}
\bibliography{my_egbib}

\end{document}